\documentclass[twoside,11pt]{article}

\usepackage[preprint]{jmlr2e}

\usepackage[utf8]{inputenc}
\usepackage[T1]{fontenc}
\usepackage{microtype}
\usepackage{graphicx}
\usepackage{booktabs}       % professional tables
\usepackage{multirow}
\usepackage{amsmath}
\usepackage{amssymb}
\usepackage{hyperref}
\usepackage{url}
\usepackage{xcolor}
\usepackage{enumitem}
\usepackage{makecell}
\usepackage{tikz}
\usepackage{flafter}       % keep floats after their source position
\usepackage{needspace}     % keep definition lead-ins with their lists

\usetikzlibrary{arrows.meta,positioning,calc,fit,backgrounds,decorations.pathreplacing}

\definecolor{rlBlue}{HTML}{2A78D6}       % Rad-Linter (M)
\definecolor{ruleOrange}{HTML}{EB6834}   % rule baseline (B1)
\definecolor{zsAqua}{HTML}{1BAF7A}       % zero-shot (Z)
\definecolor{probeYellow}{HTML}{EDA100}  % coarse probe (G)
\definecolor{inkPrimary}{HTML}{0B0B0B}
\definecolor{inkSecondary}{HTML}{52514E}
\definecolor{inkMuted}{HTML}{898781}
\definecolor{gridGray}{HTML}{E1E0D9}
\definecolor{axisGray}{HTML}{C3C2B7}
\definecolor{panelFill}{HTML}{F6F5F2}
\definecolor{errorFill}{HTML}{F9E1DD}
\definecolor{errorEdge}{HTML}{C9675C}
\definecolor{hlFill}{HTML}{FCEBB6}

\tikzset{
  fig/.style={font=\footnotesize, line cap=round, line join=round},
  flow/.style={-{Stealth[length=4.5pt,width=4pt]}, line width=0.7pt, draw=inkSecondary},
  box/.style={draw=axisGray, line width=0.6pt, rounded corners=2.5pt, fill=white,
              inner sep=4pt, align=center},
  panel/.style={draw=axisGray, line width=0.6pt, rounded corners=2.5pt, fill=panelFill},
  note/.style={font=\scriptsize, text=inkSecondary},
}

\newcommand{\fdMEasy}{98.8}
\newcommand{\fdMHard}{98.3}
\newcommand{\fdBEasy}{82.1}
\newcommand{\fdBHard}{78.9}
\newcommand{\fdZEasy}{54.5}
\newcommand{\fdZHard}{50.8}
\newcommand{\fdGEasy}{94.0}
\newcommand{\fdGHard}{56.3}
\newcommand{\fdMOmi}{97.4}
\newcommand{\fdBOmi}{98.2}
\newcommand{\fdZOmi}{77.2}
\newcommand{\fdMHed}{98.0}
\newcommand{\fdBHed}{64.7}
\newcommand{\fdZHed}{100.0}
\newcommand{\fdMCon}{98.0}
\newcommand{\fdBCon}{77.1}
\newcommand{\fdZCon}{90.8}
\newcommand{\fdMNeA}{99.5}
\newcommand{\fdBNeA}{81.7}
\newcommand{\fdZNeA}{35.1}
\newcommand{\fdMNeB}{99.2}
\newcommand{\fdBNeB}{69.5}
\newcommand{\fdZNeB}{4.6}
\newcommand{\fdNOmi}{114}
\newcommand{\fdNHed}{17}
\newcommand{\fdNCon}{131}
\newcommand{\fdNNeA}{131}
\newcommand{\fdNNeB}{131}

\hypersetup{
  pdftitle={Multimodal Error Detection in Chest X-ray Reports: A Large Language Model Reading Image Evidence},
  pdfauthor={Yanda Cheng, Bo Wen, Yuhao Lu},
  pdfsubject={Integrated paper, 2026-11-16, with supplementary material},
  pdfkeywords={radiology report quality control, image-report consistency, chest X-ray, large language models, rule-based NLP, counterfactual evaluation}
}

\newcommand{\radlinter}{\textsc{Rad-Linter}}
\newcommand{\sysM}{M}                           % reference checker: trained, scores + report
\newcommand{\bone}{B\textsubscript{1}}          % independent rule system (scores + report)
\newcommand{\sysZ}{Z}                           % zero-shot, scores + report
\newcommand{\sysT}{T}                           % trained, report only (construction check)
\newcommand{\btwo}{B\textsubscript{2}}          % zero-shot, report only (construction check)
\newcommand{\sysS}{S}                           % logistic regression, scores only (construction check)
\newcommand{\gprobe}{G}                         % pathology-agnostic coarse probe (second reference)

\ShortHeadings{Rad-Linter}{Cheng, Wen, and Lu}
\firstpageno{1}
\editor{}

\begin{document}

\title{Multimodal Error Detection in Chest X-ray Reports:\\
       A Large Language Model Reading Image Evidence}

% --- Author block: fill in for the arXiv version ---
% \author{\name Author One \email author1@institution.edu \\
%         \addr Department, Institution, City, Country
%         \AND
%         \name Author Two \email author2@institution.edu \\
%         \addr Department, Institution, City, Country}
\author{\name Yanda Cheng \email yc2675@cornell.edu \\
        \addr Meinig School of Biomedical Engineering,
              Cornell University, Ithaca, NY, USA
        \AND
        \name Bo Wen \email bwen3@ur.rochester.edu \\
        \addr Department of Electrical and Computer Engineering,
              University of Rochester, Rochester, NY, USA
        \AND
        \name Yuhao Lu \email yl3539@cornell.edu \\
        \addr School of Electrical and Computer Engineering,
              Cornell University, Ithaca, NY, USA}

\maketitle

% ============================================================
%  SECTIONS (each in its own file under sections/)
% ============================================================

% ============================================================
%  ABSTRACT  (motivation -> checker -> why a language model
%             -> controlled comparison -> results -> conclusion)
% ============================================================
\begin{abstract}
To help clinicians identify potential errors in chest X-ray
reports, we develop \radlinter{}, a language-model checker that
compares a report with evidence from the image.
This comparison is needed because a report can omit a finding or
describe one that the image does not support, without any
contradiction within the text itself.
A pretrained image classifier provides scores for nine pathologies,
and \radlinter{} flags disagreements between these scores and the
report for review.
We use a language model because this task requires interpreting
how findings are described, including negation, uncertainty, and
expressions that keyword rules may miss.
The checker is a 1.5B-parameter model fine-tuned with low-rank
adaptation (LoRA) on 1{,}816 examples.
To test whether this choice improves checking, we compare it with
a keyword-and-negation rule baseline given exactly the same reports
and classifier scores.
We evaluate both on controlled report--score pairings from OpenI,
with consistency labels derived from automatic report labels and
score thresholds.
Each test group, called a counterfactual quad, crosses two real
reports with two real score vectors from other patients to create
two consistent and two inconsistent pairs.
Every report and every score vector appears in both conditions,
so a checker using only one input can achieve only 50\% accuracy.
On 131 quads, \radlinter{} correctly classifies 98.5\% of report--score
pairs as consistent or inconsistent, compared with 80.5\% for the
rule baseline; it correctly
classifies all four pairs in 94.4\% of quads, compared with 56.5\%.
The two checkers perform similarly on omission items, which include
unmentioned and negated findings. Larger differences occur on the
remaining item types, where the checkers must interpret how a
finding is described.
The same language model without fine-tuning remains near 50\%
accuracy.
On a second set of 181 quads that also includes reports on which
the automatic labelers disagree, pair accuracy is 93.4\% for
\radlinter{} and 75.3\% for the rule baseline.
These results show that, given the same image-derived scores, a
trained language model detects report--score inconsistencies more
accurately than the tested rule baseline, with the advantage
concentrated in interpreting how findings are worded.
\end{abstract}

\begin{keywords}
radiology report quality control, image-report consistency, chest
X-ray, large language models, rule-based NLP, counterfactual
evaluation
\end{keywords}

% ============================================================
\section{Introduction}
\label{sec:intro}
% ============================================================

A chest X-ray report describes what a radiologist sees in an image
so that other clinicians can use those findings in patient care.
Errors in the report can affect later decisions, which motivates
checking the report before it is acted on~\citep{Brady2017discrepancy,
Bruno2015epidemiology,Waite2017interpretive,Zhang2023errors}.
Some errors cannot be found in the text alone. A report can leave
out a finding or describe one that the image does not support,
yet contain no contradiction between its sentences
~\citep{Kim2014perceptual,Itri2018diagnostic,Patra2021addenda,
Hanna2018fatigue,Lee2013cognitive}.
For example, a report may never mention an effusion even when there
is evidence of one in the image. To check this potential omission,
a reviewer needs evidence about the image as well as the report.

Text-based tools can help with errors that leave clues in the
report. Keyword and template rules identify missing or inconsistent
statements~\citep{Donnelly2022nlp,Chng2023labelling}; neural systems
support report review and classify the findings described
~\citep{Min2022rred,Banerjee2023nlp,Pan2025radclare}.
Language models have also been tested on errors inserted into
report text~\citep{Gertz2024gpt4,Salam2025closed_open,Sun2025llm_error}.
A measurement typo or a disagreement between the detailed findings
and the report's summary may be detectable this way. Checking a
statement against the image requires an additional source of
evidence.

We develop \radlinter{} to make that comparison. First, an image
classifier, a model trained to recognize abnormalities in chest
X-rays, produces one score for each of nine findings, or pathologies
~\citep{Rajpurkar2017chexnet,Irvin2019chexpert,Cohen2022xrv}.
A higher score indicates stronger support from that classifier for
the finding. A language model then receives the report and these
scores and flags potential disagreements. A high effusion
score paired with a report that does not affirm effusion is one
example. Multimodal checkers can instead receive the image and
report directly~\citep{Zou2025corbenchx,Wu2024multimodal}.
Here, the language model receives scores rather than image pixels.
This makes the evidence supplied to it explicit and lets us test
its response when that evidence changes. Our clinical aim is to
identify problems worth reviewing. The task evaluated here is to
detect disagreement between the report and the supplied scores.
Either source may be wrong, and the checker may misread the report;
a clinician must review a flag before treating it as a report error.

We use a language model because the comparison depends on how a
finding is described. The same pathology name may occur in
``effusion is present,'' ``no effusion,'' or ``possible effusion.''
A keyword match alone does not distinguish those meanings. Rules
can add checks for negation and uncertainty, but the relevant words
may be far apart. In ``no evidence of consolidation, pneumothorax,
or pleural effusion,'' the negation applies to all three findings.
A report may also express uncertainty in one sentence and affirm
the finding in another. We therefore fine-tune Qwen2.5-1.5B, a
language model with 1.5 billion learned parameters, on this comparison
task.
Fine-tuning means further training an existing model on examples of
the desired input and answer. We use low-rank adaptation
(LoRA)~\citep{hu2022lora}, which trains a small set of added weights
while leaving the model's original weights fixed.

To test this choice, we compare the trained model with a rule
baseline: a comparison checker that uses keywords, negation cues,
and uncertainty cues. Both receive exactly the same reports and
scores and return the same decision format. This holds the image
evidence fixed while changing how the report is read. Each system
must decide whether there is an inconsistency and, if so, name its
type and the affected finding.

A fair test must also make the evidence necessary. In an earlier
version with errors inserted into report text, a model given only
the report scored well (\S\ref{sec:method:test}). That result could
not show whether a checker compared the report with the scores.
We instead keep the reports unchanged and form
\emph{counterfactual quads}, groups of four test pairs. Each of two
reports is paired with each of two score vectors, where a vector is
simply a list of nine scores. Every report and every vector appears
once in a consistent pair and once in an inconsistent pair. A
checker using only one input must give the same answer to two pairs
with opposite reference labels, so it can be correct on only half.
Additional tests examine whether a checker uses broad clues, such
as the presence of any high score, without matching individual
findings.

We build these tests from OpenI, a public collection of chest
X-rays and reports~\citep{Demner2016openi}. A rule-based reference
labeler assigns report states, which are combined with score
thresholds to generate the test answers. An agreement labeler,
implemented with another language model, provides a second reading
to select primary-test reports on which the readings agree.
On 131 quads, \radlinter{} correctly classifies 98.5\% of individual
pairs as consistent or inconsistent, compared with 80.5\% for the
rule baseline.
The two perform similarly on the \textsc{Omission} category,
which includes both unmentioned and negated findings paired with
high scores. Larger differences occur in the other item types,
which require interpreting how a finding is stated.
The same language model without task-specific training remains
near 50\% accuracy. On a second test set that includes reports on
which the automatic labelers disagree, the trained model retains
an advantage, with 93.4\% accuracy against 75.3\%.
These results motivate examining the language component of report
checking while holding the image evidence constant. Prior work on
report review, image evidence, and test design provides the basis
for this comparison.

% ============================================================
\section{Related Work}
\label{sec:related}
% ============================================================

\paragraph{Reading the report.}
Text-based report review uses language to find problems within a
report. Rule systems match keywords and
patterns~\citep{Donnelly2022nlp,Chng2023labelling}, while neural
systems support error detection and finding
classification~\citep{Min2022rred,Banerjee2023nlp}.
RadCLARE applies automated review across an
institution~\citep{Pan2025radclare,Devlin2019bert}.
\citet{Casey2021nlpreview} review natural language processing,
meaning automated analysis of language, in radiology;
\citet{Nobel2022structured} review structured reporting, which uses
predefined report fields. Language models offer another way to
read reports. \citet{Gertz2024gpt4} inserted 150 textual errors into
100 of 200 reports and found GPT-4 close to radiologists at detecting
them. Later studies evaluated closed and open
models~\citep{Salam2025closed_open,Sun2025llm_error} and errors from
speech recognition~\citep{Schmidt2024speech}.
These studies support automated text review. Our task additionally
asks whether a statement agrees with evidence outside the report.

\paragraph{Supplying image evidence.}
Image-based approaches give a checker information that the report
alone cannot provide. CorBenchX~\citep{Zou2025corbenchx} evaluates
vision-language models, which process images and text together,
on reports with five types of inserted error. Its source is
MIMIC-CXR, a public collection of chest X-rays and reports.
\citet{Wu2024multimodal} fine-tune a model based on the vision-language
model LLaVA using MIMIC-CXR, test it on IU X-ray (the OpenI collection
used here), and compare it with clinicians.
Other work represents the image through classifier outputs.
\citet{Mahmood2023factcheck,Mahmood2025phrase} check generated reports
against images, \citet{Warr2024quality} use disease classifiers to
audit generated reports, and ChatCAD puts classifier outputs into
a language-model prompt as text~\citep{Wang2023chatcad}.
We use a score-based interface to ask which checker compares the
same evidence with the report more accurately.

Related research evaluates the accuracy of automatically generated
reports. Factuality metrics measure whether their statements are
supported~\citep{Yu2023evaluating,Nguyen2023pragmatic,
Miura2021factual,Delbrouck2022semantic,Ostmeier2024green}.
Grounded generation links report statements to image evidence, as
in MAIRA-2~\citep{Bannur2024maira2} and
MAIRA-Seg~\citep{Sharma2024maira}. RadFlag checks for unsupported
content~\citep{Zhang2024radflag}, and \citet{Moll2025faithfulness}
test whether medical vision-language models base their responses
on the supplied input.
These studies share the goal of connecting statements to evidence.
Our comparison fixes the reports and scores and changes the method
used to check their agreement.

\paragraph{Testing whether the evidence is used.}
Providing two inputs does not ensure that a model uses both.
In natural language inference, a model judges whether one sentence
follows from another. Models can score well even when shown only
the second sentence~\citep{Gururangan2018hypothesis}.
This is an example of a shortcut: a clue predicts the test answer
without requiring the intended reasoning. Shortcuts have been
studied across deep learning~\citep{Geirhos2020shortcut} and in
chest X-ray classification~\citep{Zech2018cxr,DeGrave2021shortcut}.
Balanced test pairs help expose them. VQA~v2 uses complementary
images for visual questions~\citep{Goyal2017vqa2}; Winoground uses
four image--caption pairings from two images and two
captions~\citep{Thrush2022winoground}; and symmetric FEVER pairs
balance claims and supporting evidence~\citep{Schuster2019symmetric}.
Contrast sets~\citep{Gardner2020contrast} and counterfactually
augmented data~\citep{Kaushik2020cad} likewise change selected
parts of examples to test whether predictions change appropriately.

Our task follows the claim--evidence approach of
FEVER~\citep{Thorne2018fever} and its scientific and health
applications~\citep{Wadden2020scifact,Kotonya2020pubhealth,
Sarrouti2021healthver,Guo2022factcheck_survey}.
Radiology benchmarks such as ReXErr~\citep{Rao2024rexerr} and
CorBenchX create errors by editing reports; synthetic clinical
text is also used in training~\citep{Tang2023synthetic,
Liu2025synthetic_fracture,Li2023twodirections}.
We instead vary the scores paired with unchanged reports.
Following \citet{Hessel2020emap}, we also test whether simple
summaries of both inputs can explain performance. These ideas lead
to two requirements for our evaluation: each input alone must be
insufficient, and success must be examined beyond a few broad
clues. The method below defines the inconsistency that the checker
must detect, then builds a comparison that meets these requirements.

\section{Method}
\label{sec:method}

\subsection{What the Checker Detects}
\label{sec:problem}
\label{sec:method:problem}

The checker receives one report and a list of image-classifier scores
for nine findings. It returns one judgment: whether the pair is
inconsistent and, if so, the type of disagreement and the affected
finding. The test pairs are constructed so that at most one finding
is inconsistent. We evaluate this comparison with the supplied scores;
the test does not determine whether a report is clinically wrong.

\Needspace{7\baselineskip}
Under the test definition, a high score calls for an affirmed finding
in the report, while a low score conflicts with an affirmed finding.
For example, a high effusion score paired with ``no effusion'' is
inconsistent. Correctly flagging that disagreement counts as correct
in this test even if a clinician later finds that the classifier was
wrong. A checker can also misread a report, so a flag itself still
requires review.

The output labels describe these score--report combinations.
\textsc{Omission} means a high score paired with an unmentioned or
explicitly negated finding; \textsc{Hedging} means a high score paired
with uncertain wording. \textsc{Contradiction} means a low score paired
with an affirmed finding. \textsc{No\_Error} means that none of these
conditions is detected across the nine findings. Uncertainty alone
is not an error: its treatment here is a test convention, not a
clinical rule. Appendix~\ref{supp:task} gives the complete state table,
thresholds, and output fields.

\subsection{Building and Training \radlinter{}}
\label{sec:method:system}

\begin{figure}[!htb]
\centering
% V06: restore evidence and output meaning; distinguish reference from predictions.
\begin{tikzpicture}[fig,x=1pt,y=1pt]
\path[use as bounding box] (0,0) rectangle (432,-341);
\node[panel,anchor=north west,minimum width=242pt,minimum height=181pt] at (0,0) {};
\node[panel,anchor=north west,minimum width=178pt,minimum height=181pt] at (254,0) {};
\node[anchor=north west,font=\small\bfseries] at (6,-5) {Image-derived scores};
\node[anchor=north west,font=\small\bfseries] at (260,-5) {Report excerpt};
\begin{scope}[shift={(10,-52)}]
 \fill[inkPrimary!90,rounded corners=2pt] (0,0) rectangle (20,26);
 \fill[inkMuted!75] (10,12) ellipse (8 and 11);
 \fill[inkPrimary!78] (6,14) ellipse (3 and 8);
 \fill[inkPrimary!78] (14,14) ellipse (3 and 8);
\end{scope}
\node[box,minimum width=180pt,minimum height=26pt,font=\footnotesize] (clf) at (140,-39) {Fixed image classifier};
\draw[flow] (30,-39) -- (clf.west);
\node[anchor=west,font=\fontsize{8.5}{10}\selectfont,inner sep=0pt,text=inkSecondary] at (10,-70) {Nine scores};
\node[anchor=south,font=\fontsize{8.5}{10}\selectfont,align=center,inner sep=1pt] at (144.6,-76) {Low\\0.2};
\node[anchor=south,font=\fontsize{8.5}{10}\selectfont,align=center,inner sep=1pt] at (173.8,-76) {High\\0.6};
\foreach \name/\s [count=\i] in {Atelectasis/0.71,Consolidation/0.12,Pneumothorax/0.05,Edema/0.02,Effusion/0.08,Pneumonia/0.15,Cardiomegaly/0.33,Nodule/0.41,Mass/0.18} {
 \pgfmathsetmacro{\yy}{-85-10*(\i-1)}
 \node[anchor=east,font=\fontsize{8.5}{10}\selectfont,inner sep=0pt] at (124,\yy) {\name};
 \fill[gridGray] (130,\yy-2) rectangle (203,\yy+2);
 \ifnum\i=1
  \fill[inkPrimary] (130,\yy-2) rectangle (130+73*\s,\yy+2);
  \node[anchor=west,font=\footnotesize,inner sep=2pt] at (207,\yy) {0.71};
 \else
  \fill[inkMuted] (130,\yy-2) rectangle (130+73*\s,\yy+2);
 \fi
}
\draw[inkSecondary,line width=.5pt] (144.6,-78) -- (144.6,-170);
\draw[inkSecondary,line width=.5pt] (173.8,-78) -- (173.8,-170);
\node[anchor=north west,text width=160pt,align=left,font=\footnotesize,inner sep=0pt] at (263,-31)
 {\hyphenpenalty=10000\raggedright\textsc{Findings}: Heart size is normal. Bibasilar opacity,\\
 \colorbox{hlFill}{may represent atelectasis.}\\
 No pleural effusion or pneumothorax.\\[8pt]
 \textsc{Impression}:\\\colorbox{hlFill}{Possible} bibasilar atelectasis.};
\draw[inkSecondary,line width=.7pt] (121,-181) -- (121,-198) -- (343,-198) -- (343,-181);
\node[fill=white,inner sep=2pt,font=\footnotesize] at (232,-198) {Same report and scores};
\draw[flow] (232,-202) -- (232,-210) -| (104,-224);
\draw[flow] (232,-210) -| (328,-224);
\node[box,anchor=north,minimum width=208pt,minimum height=28pt,draw=rlBlue,line width=1pt,font=\small\bfseries] at (104,-224) {\radlinter{}};
\node[box,anchor=north,minimum width=208pt,minimum height=28pt,draw=ruleOrange,line width=1pt,font=\small\bfseries] at (328,-224) {Rule baseline};
\draw[flow] (104,-252) -- (104,-266);
\draw[flow] (328,-252) -- (328,-266);
\node[box,anchor=north,minimum width=120pt,minimum height=22pt] at (104,-266) {Prediction};
\node[box,anchor=north,minimum width=120pt,minimum height=22pt] at (328,-266) {Prediction};
\node[anchor=north west,font=\footnotesize,inner sep=0pt] at (0,-299)
 {\textbf{Output fields:} inconsistency decision, type, finding};
\node[panel,anchor=north west,minimum width=432pt,minimum height=25pt] at (0,-316) {};
\node[anchor=west,font=\footnotesize,inner sep=0pt] at (7,-328)
 {\textbf{Example reference:} Inconsistent\enspace|\enspace Hedging\enspace|\enspace Atelectasis};
\end{tikzpicture}
\caption{Rad-Linter and the rule baseline receive the same report
and nine image-classifier scores, then independently predict the
inconsistency decision, type, and finding. The illustrative example
pairs a high Atelectasis score with uncertain wording, giving a
\textsc{Hedging} reference label; this is not a measured prediction
from either checker. High and low scores use cut-offs of 0.6 and
0.2, respectively; the scores are not validated clinical probabilities.}
\label{fig:overview}
\end{figure}
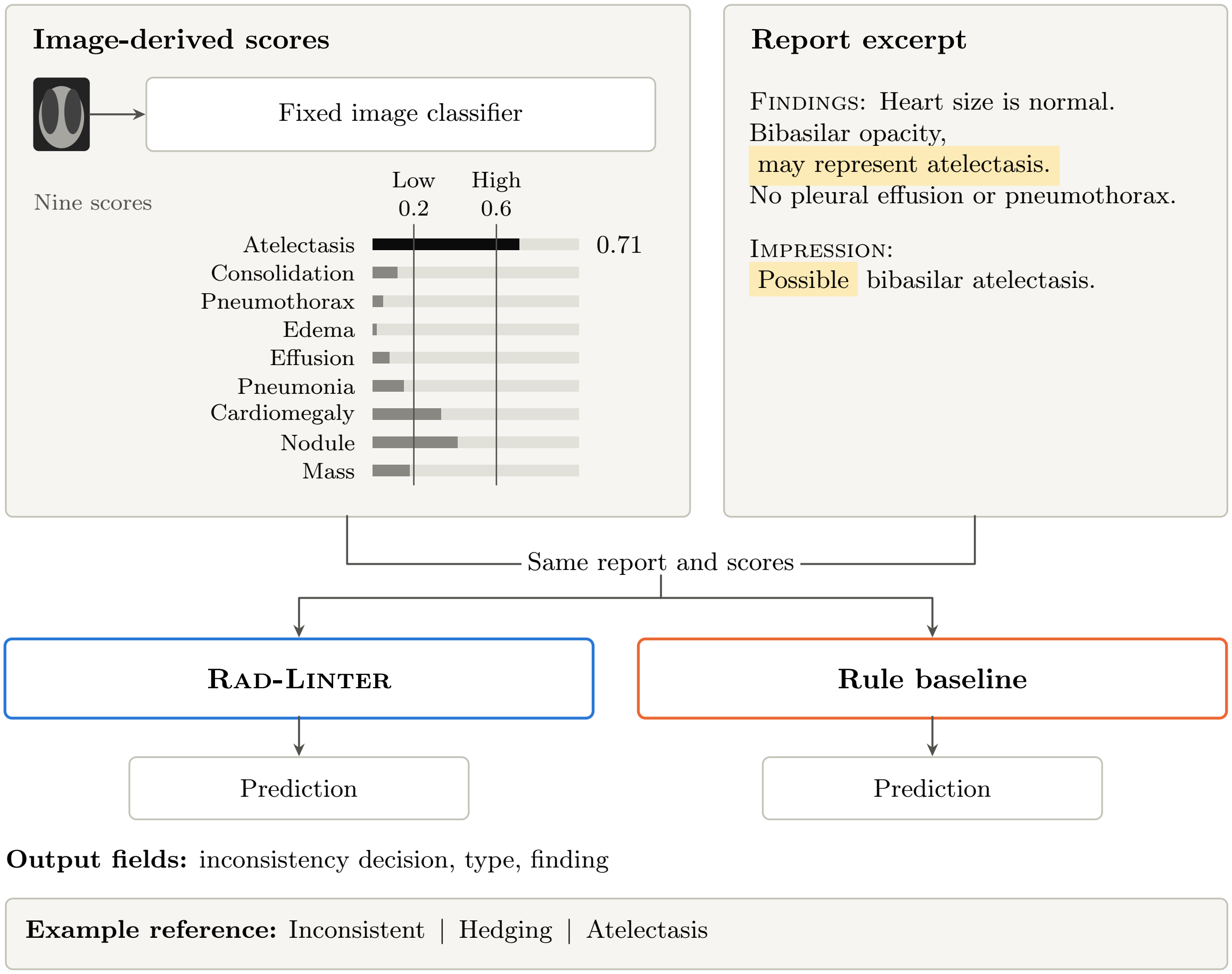

\radlinter{} has two stages (Figure~\ref{fig:overview}). A fixed
image classifier supplies the scores; a fine-tuned language model
reads them with the unchanged report. We use TorchXRayVision's
DenseNet-121 classifier~\citep{Cohen2022xrv,Huang2017densenet}, retaining
nine of its 18 outputs without retraining it. Thus, the language
model receives explicit scores rather than image pixels. The rule
baseline receives exactly the same report and scores.

We fine-tune Qwen2.5-1.5B-Instruct~\citep{qwen25} using low-rank
adaptation (LoRA)~\citep{hu2022lora}, which updates small added modules
while keeping the original model weights fixed. Training uses
1{,}816 labeled report--score pairs in 454 quads, the four-pair groups
described in \S\ref{sec:method:test}. It teaches the model to produce
the three output fields for each pair. We train for 30 epochs, or
30 passes through these examples, and repeat training with three
random seeds. The 1.5B-parameter model fits on one consumer GPU.

The selected recipe followed an earlier recipe that fit the training
examples poorly. Revised recipes were compared on a development
split after the initial recipe had already been evaluated on the
test sets; \S\ref{sec:limitations} explains this limitation.
Appendix~\ref{supp:training} retains the classifier's training sources,
model settings, exact prompts, and recipe history. The final training
reports were not filtered for agreement between labelers.

\subsection{Comparing with the Rule Baseline}
\label{sec:method:rules}

The \emph{rule baseline} is the checker against which we compare
Rad-Linter. It matches pathology keywords and nearby negation or
uncertainty cues, following the local cue-based approach of
NegEx~\citep{Chapman2001negex}. It uses the same score thresholds
and output fields as Rad-Linter. The comparison therefore holds
the supplied evidence and decision task constant while changing
how the report is interpreted.

These baseline rules differ from the rules used to prepare reference
answers. We call the latter the \emph{reference labeler}: it supplies
report states for data preparation, rather than serving as the rule
baseline in the comparison. Appendix~\ref{supp:labeling} documents
both implementations. Their distinct roles matter when interpreting
a result in which the language model outperforms the tested baseline.

\subsection{Constructing a Test That Requires Both Inputs}
\label{sec:method:test}

\paragraph{Reports and reference answers.}
We use the 3{,}599 chest X-ray reports in OpenI~\citep{Demner2016openi},
one per patient. Each report combines its Findings and Impression
sections and has up to four frontal images, each with a score vector.
We split patients into 2{,}520 for training and 1{,}079 held out for
testing. All pairings are constructed within their respective splits.

The reference labeler uses rules to assign each finding a report
state. These states, combined with score thresholds, produce the
reference answers used for training and evaluation. A separate
\emph{agreement labeler}, a language model, provides another reading
of the report under the same conventions. Agreement between the
two labelers is required for primary-test reports, which must also
meet the pairing constraints below. The agreement labeler does not
grade either checker's output. Agreement is a selection criterion,
not clinical validation of the answers. Exact rules, the three-way
state comparison, and the agreement prompt are in
Appendix~\ref{supp:labeling}.

\paragraph{Four pairings with different answers.}
In an earlier evaluation with errors inserted into report text,
systems could score well using only one input
(Appendix~\ref{supp:evaluation}). We instead keep reports unchanged
and vary the evidence paired with them. For one target finding,
report $A$ does not affirm it, while report $B$ does. We select a
high-score vector and a low-score vector from other patients in the
same split, then pair each report with each vector. Both vectors
must agree with both reports on the other eight findings, so only
the target can disagree.

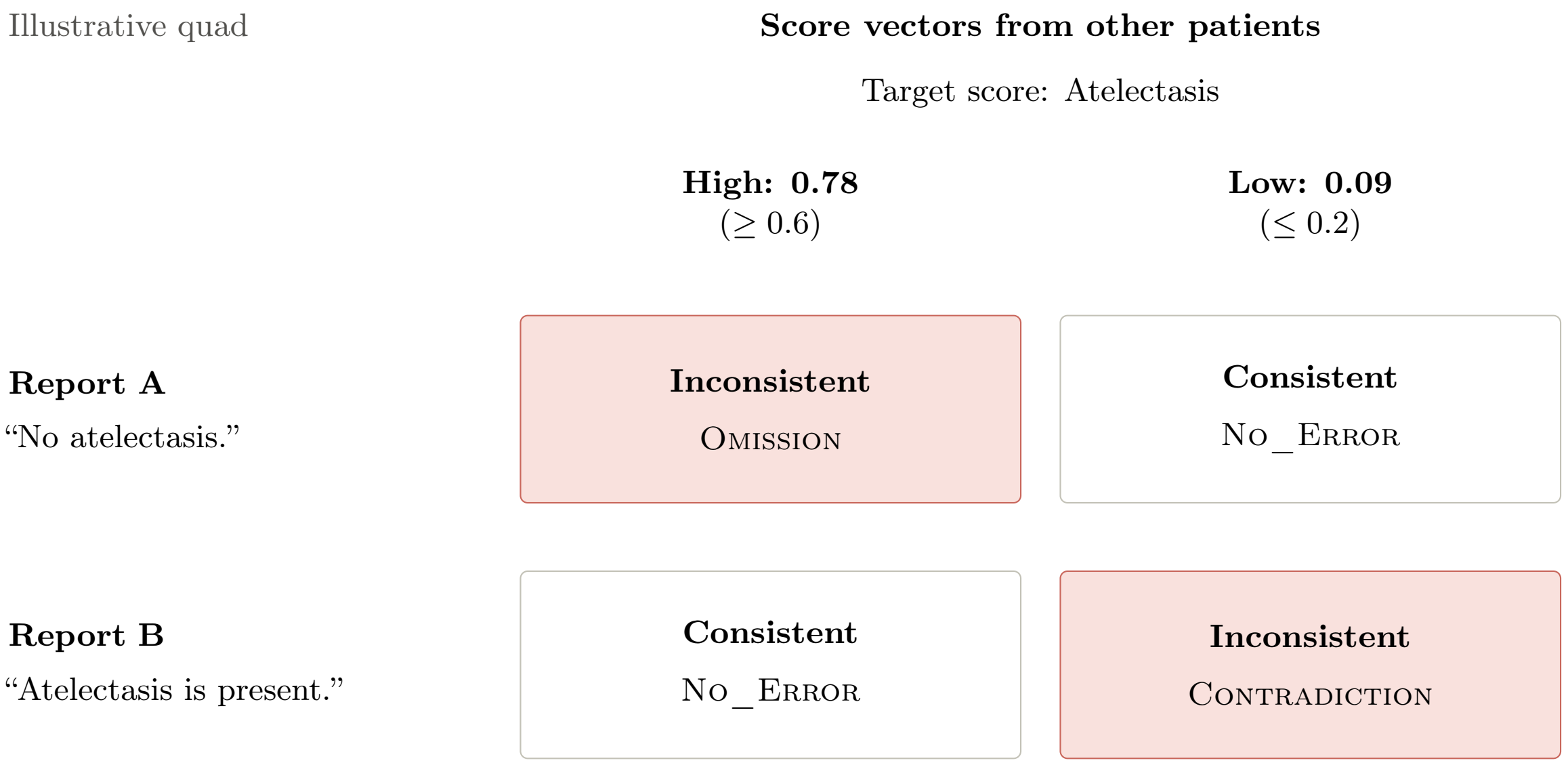
\begin{figure}[!htb]
\centering
% V06: concrete quad, complete score-vector identity and task-category mapping.
\begin{tikzpicture}[fig,x=1pt,y=1pt,
 cell/.style={draw=axisGray,rounded corners=2pt,fill=white,minimum width=139pt,
 minimum height=52pt,align=center,font=\footnotesize,inner sep=4pt},
 mismatch/.style={cell,fill=errorFill,draw=errorEdge}]
\path[use as bounding box] (0,0) rectangle (432,-213);
\node[anchor=north west,font=\footnotesize,text=inkSecondary,inner sep=0pt] at (0,0) {Illustrative quad};
\node[anchor=north,font=\footnotesize\bfseries,inner sep=0pt] at (287,0) {Score vectors from other patients};
\node[anchor=north,font=\footnotesize,inner sep=0pt] at (287,-18) {Target score: Atelectasis};
\node[anchor=north,align=center,font=\footnotesize] at (212,-40) {\textbf{High: 0.78}\\($\geq 0.6$)};
\node[anchor=north,align=center,font=\footnotesize] at (362,-40) {\textbf{Low: 0.09}\\($\leq 0.2$)};
\node[anchor=west,text width=127pt,align=left,font=\footnotesize,inner sep=0pt] at (0,-110)
 {\textbf{Report A}\\[4pt]``No atelectasis.''};
\node[anchor=west,text width=127pt,align=left,font=\footnotesize,inner sep=0pt] at (0,-181)
 {\textbf{Report B}\\[4pt]``Atelectasis is present.''};
\node[mismatch] at (212,-110) {\textbf{Inconsistent}\\[5pt]\textsc{Omission}};
\node[cell] at (362,-110) {\textbf{Consistent}\\[5pt]\textsc{No\_Error}};
\node[cell] at (212,-181) {\textbf{Consistent}\\[5pt]\textsc{No\_Error}};
\node[mismatch] at (362,-181) {\textbf{Inconsistent}\\[5pt]\textsc{Contradiction}};
\end{tikzpicture}
\caption{An illustrative counterfactual quad for Atelectasis.
Each report is paired with each full score vector from other
patients; only the target score is displayed. The other eight
findings agree in all four pairs. Negation in report $A$ produces
an \textsc{Omission} label at the high score; an uncertain $A$
would instead produce \textsc{Hedging}. A fixed valid binary
decision based on either input alone is correct on two of the
four pairs.}
\label{fig:quad}
\end{figure}

For example, let $A$ say ``no atelectasis'' and $B$ say ``atelectasis
is present.'' The high score disagrees with $A$ and agrees with $B$;
the low score agrees with $A$ and disagrees with $B$. The group has
two consistent and two inconsistent pairs. Each pair is an
\emph{item}; the group of four is a \emph{counterfactual quad}
(Figure~\ref{fig:quad}). The reports and scores are real, but these
cross-patient pairings are constructed for the experiment.

A checker seeing only report $A$ must give the same answer for both
scores, although the reference answers differ. One answer is then
correct and one incorrect. The same reasoning applies to $B$ and
to a checker seeing only a score vector. This makes either input
alone insufficient. It does not rule out every shortcut that uses
both inputs, which motivates the additional checks below.

\paragraph{Primary and second tests.}
The primary test has 131 quads, or 524 items, formed from held-out
reports that meet the agreement filter. A second test is constructed
from the same held-out patient split without requiring labeler
agreement. It has 181 quads, or 724 items, with reference answers
still generated from the reference labeler's states and score
thresholds. It is constructed separately, not by simply adding
cases to the primary test. It tests whether the comparison still
favors Rad-Linter without the agreement filter. Dataset composition
and input-reuse limits are retained in Appendix~\ref{supp:evaluation}.

\subsection{Additional Checks and Performance Measures}
\label{sec:method:eval}

Our main comparison is Rad-Linter versus the rule baseline. Three
additional checks examine possible explanations for the results.
First, \emph{single-input checks} receive only the report or only
the scores. They verify the expected behavior of the balanced pairs.
Second, a \emph{coarse probe} uses broad summaries of both inputs,
but cannot match a particular finding to its score. Third, the
\emph{zero-shot model} is the same base language model given both
inputs without task-specific fine-tuning. It tests what this model
can achieve from the task description alone. Complete system
identities and settings are in Appendix~\ref{supp:evaluation}.

The coarse probe addresses a specific shortcut: when report $A$
affirms no finding, ``nothing affirmed'' together with ``some score
is high'' can often identify an inconsistency without matching
individual findings. In 64 of the primary test's 131 quads, report
$A$ affirms another finding. We call this the \emph{hard subset},
because that simple cue is less useful. It is a subset of the
primary test, not the second, unfiltered test. The names hard and
easy describe this shortcut, not clinical difficulty. The probe
uses states from the reference labeler and summaries of the scores;
its seven features are specified in Appendix~\ref{supp:evaluation}.

\emph{Item accuracy} is the fraction of pairs correctly classified
as consistent or inconsistent. \emph{Quad accuracy} is the fraction
of groups for which all four such decisions are correct. Three
correct pairs contribute to item accuracy but do not solve a quad.
We also assess missed inconsistencies, false flags, and the predicted
finding and category; these measures are defined in
Appendix~\ref{supp:evaluation}.

Results for repeated training runs give the mean and standard
deviation across three seeds. Square brackets indicate 95\%
bootstrap intervals where reported. To allow for reused reports or
score vectors, the bootstrap resamples groups of related quads
together, using the same sampled groups when comparing systems.
These intervals measure uncertainty within this test, while the
seed variation describes repeated training. The complete calculation
and parsing rules are in Appendix~\ref{supp:evaluation}.

\section{Results}
\label{sec:experiments}

\subsection{Comparing Rad-Linter with the Rule Baseline}
\label{sec:exp:find}
\label{sec:exp:rules}

On the primary controlled test, Rad-Linter correctly classifies
98.5\% of report--score pairs as consistent or inconsistent,
compared with 80.5\% for the rule baseline. Correctness is measured
against the automatically generated consistency labels. These are
not rates of detecting clinically confirmed report errors.
Rad-Linter correctly answers all four items in 94.4\% of quads,
compared with 56.5\% for the baseline (Table~\ref{tab:main}).
The model's results are means over three training runs.

\begin{table}[!htb]
\centering
\caption{Main comparison on the two tests. Rad-Linter entries give
mean $\pm$ standard deviation across three training seeds; brackets
for the deterministic rule baseline give 95\% cluster-bootstrap
intervals. These describe different sources of variation.
Item accuracy concerns individual binary decisions; quad accuracy
requires all four decisions to be correct. Complete system results
are retained in Appendix~\ref{supp:evaluation}.}
\label{tab:main}
\begin{tabular}{lcc}
\toprule
Checker & Item accuracy (\%) & Quad accuracy (\%) \\
\midrule
\multicolumn{3}{l}{\textit{Primary test: 131 quads, 524 items}} \\
Rad-Linter & $98.5\pm0.2$ & $94.4\pm0.4$ \\
Rule baseline & 80.5 [75.0, 86.6] & 56.5 [36.7, 64.3] \\
\midrule
\multicolumn{3}{l}{\textit{Without agreement filter: 181 quads, 724 items}} \\
Rad-Linter & $93.4\pm0.6$ & $81.4\pm2.1$ \\
Rule baseline & 75.3 [73.1, 82.9] & 47.0 [37.2, 57.5] \\
\bottomrule
\end{tabular}
\end{table}

The advantage is about 18 percentage points on items and 38 on
quads, where percentage points are differences between the two
accuracy percentages. It appears in all three training runs.
Paired bootstrap intervals for each run's item difference are
$[+13.4,+25.0]$ points; the quad intervals lie within
$[+33.1,+63.3]$. Their positive lower bounds indicate that the
advantage persists when related test cases are resampled together.
This measures uncertainty within the test, not performance in a
new clinical population.

The model both misses fewer inconsistencies and raises fewer false
flags. It flags 97.7\% of inconsistent items, versus 85.5\% for the
baseline, and leaves 99.4\% of consistent items unflagged, versus
75.6\%. Its false flags therefore affect 0.6\% of consistent items,
or one to two per run. The advantage also remains when tested
pathologies receive equal weight: 94.5\% mean accuracy for the
model and 81.5\% for the baseline. Results for individual findings
vary, with small samples limiting several estimates; all such
results are retained in Appendix~\ref{supp:evaluation}.

\begin{figure}[!htb]
\centering
% V06: main-system differences shown by paired markers; zero-shot offset below.
\begin{tikzpicture}[fig,x=1pt,y=1pt]
\path[use as bounding box] (0,0) rectangle (432,-261);
\node[anchor=north west,font=\footnotesize\bfseries,inner sep=0pt] at (0,0) {Primary test (524 pairs)};
\fill[rlBlue] (4,-23) circle (2.8pt);
\node[anchor=west,font=\footnotesize,inner sep=2pt] at (11,-23) {\radlinter{} (3-run mean)};
\draw[ruleOrange,line width=1pt,fill=white] (210,-26.5) rectangle (217,-19.5);
\node[anchor=west,font=\footnotesize,inner sep=2pt] at (222,-23) {Rule baseline};
\draw[inkSecondary,line width=.8pt,fill=white] (350,-19) -- (346,-26) -- (354,-26) -- cycle;
\node[anchor=west,font=\footnotesize,inner sep=2pt] at (358,-23) {Zero-shot};
\def\xo{176}\def\k{1.85}
\foreach \t in {0,20,...,100} {
 \draw[gridGray,line width=.5pt] (\xo+\t*\k,-65) -- (\xo+\t*\k,-229);
 \node[anchor=north,font=\scriptsize,inner sep=2pt] at (\xo+\t*\k,-231) {\t};
}
\node[anchor=south,align=center,text width=56pt,font=\fontsize{8.5}{10}\selectfont,inner sep=0pt] at (404,-62)
 {\radlinter{}\\$-$ rule\\(points)};
\node[anchor=west,font=\footnotesize\bfseries,inner sep=0pt] at (0,-53) {Inconsistent pairs};
\node[anchor=west,font=\footnotesize\bfseries,inner sep=0pt] at (0,-165) {Consistent pairs (\textsc{No\_Error})};
\foreach \desc/\detail/\m/\b/\z/\yy in {
 {Omission}/{Unmentioned/denied; high score; $n=\fdNOmi$}/\fdMOmi/\fdBOmi/\fdZOmi/-74,
 {Hedging}/{Uncertain; high score; $n=\fdNHed$}/\fdMHed/\fdBHed/\fdZHed/-105,
 {Contradiction}/{Affirmed; low score; $n=\fdNCon$}/\fdMCon/\fdBCon/\fdZCon/-136,
 {Not affirmed}/{Low score; $n=\fdNNeA$}/\fdMNeA/\fdBNeA/\fdZNeA/-187,
 {Affirmed}/{High score; $n=\fdNNeB$}/\fdMNeB/\fdBNeB/\fdZNeB/-218} {
 \node[anchor=west,font=\footnotesize,inner sep=0pt] at (0,\yy+3) {\desc};
 \node[anchor=west,font=\fontsize{8.5}{10}\selectfont,text=inkSecondary,inner sep=0pt] at (0,\yy-9) {\detail};
 \draw[axisGray,line width=1.3pt] (\xo+\b*\k,\yy) -- (\xo+\m*\k,\yy);
 \draw[ruleOrange,line width=1pt,fill=white]
  (\xo+\b*\k-3.5,\yy-3.5) rectangle (\xo+\b*\k+3.5,\yy+3.5);
 \fill[rlBlue] (\xo+\m*\k,\yy) circle (2.8pt);
 \draw[inkSecondary,line width=.8pt,fill=white]
  (\xo+\z*\k,\yy-5) -- (\xo+\z*\k-3.5,\yy-11) -- (\xo+\z*\k+3.5,\yy-11) -- cycle;
 \pgfmathsetmacro{\gap}{\m-\b}
 \node[font=\footnotesize,inner sep=0pt] at (404,\yy)
  {\pgfmathprintnumber[fixed,fixed zerofill,precision=1,showpos]{\gap}};
}
\node[anchor=north,font=\footnotesize,inner sep=0pt] at (269,-249) {Consistency accuracy (\%)};
\end{tikzpicture}
\caption{Binary consistency accuracy by reference item type on
the primary test, not accuracy of predicting the type itself.
Each row specifies the score--report condition and sample size.
The two consistent groups share \textsc{No\_Error}. Connecting
segments compare Rad-Linter with the rule baseline; the right
column gives their difference in percentage points. Zero-shot
markers are offset vertically for visibility. Rad-Linter values
average three training seeds; the other systems have single
deterministic runs.}
\label{fig:by_type}
\end{figure}
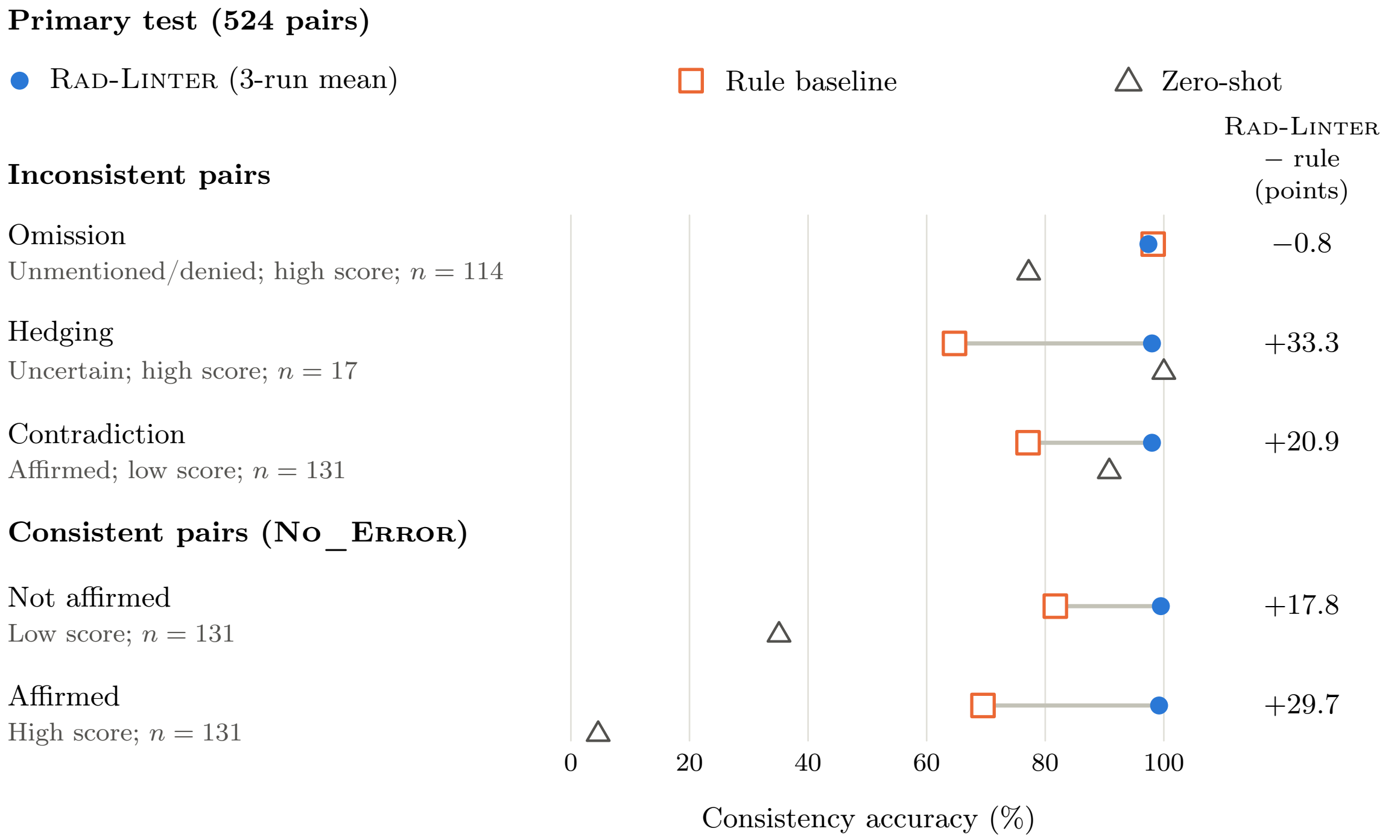

\Needspace{6\baselineskip}
\paragraph{Where the difference is larger.}
The systems perform similarly when a high score is paired with an
unmentioned or explicitly negated finding, the \textsc{Omission}
category: 97.4\% for Rad-Linter and 98.2\% for the baseline.
Larger differences occur in the remaining item groups
(Figure~\ref{fig:by_type}). The baseline scores 64.7\% on high-score
pairs with uncertain wording (\textsc{Hedging}), 77.1\% on low-score
pairs with affirmed findings (\textsc{Contradiction}), and 69.5\% on
consistent high-score pairs with affirmed findings.

This pattern is consistent with more reliable interpretation of
report wording under the reference-labeling conventions. It does
not isolate a particular linguistic mechanism: each category mixes
language phenomena, and we have not counted errors using an
independently annotated linguistic analysis. Because both checkers
receive the same scores, access to a better image classifier cannot
explain the difference.

\subsection{Checking Alternative Explanations}
\label{sec:exp:checks}

\paragraph{Could either input alone explain the result?}
The single-input checks, receiving only reports or only scores,
all reach 50\% item accuracy and 0\% quad accuracy. This is the
expected result of the balanced pairings: a fixed valid binary
answer cannot distinguish the two labels attached to the same
input. It confirms the construction check, rather than showing
that these systems lack all useful knowledge.

\paragraph{Could broad clues from both inputs explain it?}
The coarse probe reaches 75.6\% item accuracy overall and 94.0\%
on the 268 easy items, where report $A$ affirms nothing. Thus,
noticing ``nothing affirmed'' and ``some score is high'' can often
work without matching findings to scores. On the 256 items in the
hard subset, where $A$ affirms another finding, the probe falls to
56.3\%, while Rad-Linter reaches 98.3\%, close to its 98.8\% on
easy items. These specified broad features do not explain the
model's performance; this check does not exclude every possible
shortcut. The probe uses an earlier training set and has no
reported confidence interval (\S\ref{sec:limitations}); its full
comparison is in Appendix~\ref{supp:evaluation},
Figure~\ref{fig:hard_easy}.

\paragraph{Does task-specific training matter for this model?}
The zero-shot model, the same generally pretrained model without
this task's fine-tuning, reaches only 52.7\% item accuracy and
solves no quad. It flags 224 of 262 inconsistent items but also
210 of 262 consistent ones. Training on the 1{,}816 examples
produces much better separation of these cases. This establishes
the value of training in this setup, not that every language model
would require the same training. Exact prompts and output-target
settings are preserved in Appendix~\ref{supp:training}.

\subsection{Testing Without the Agreement Filter}
\label{sec:exp:all}

The second test asks whether the advantage persists when reports
are not selected for agreement between the reference labeler and
agreement labeler. It is constructed separately from the same
held-out patient split and still uses automatic reference answers.
It is not the hard subset of the primary test and is not an
external clinical validation.

Rad-Linter reaches 93.4\% item accuracy and 81.4\% quad accuracy,
compared with 75.3\% and 47.0\% for the rule baseline
(Table~\ref{tab:main}). The differences are about 18 percentage
points on items and 34 on quads. The zero-shot model remains near
chance at 50.8\% item accuracy. Complete results and variation
across runs are in Appendix~\ref{supp:evaluation}.

Both checkers score lower than on the primary test. Report
selection and the composition of the pairs differ, including more
items with uncertain wording,
so this comparison does not isolate the cause of the decrease.
It does show that Rad-Linter's advantage is not confined to reports
on which the labelers agree. Together, the main comparison and
these checks support a trained language model as the report-reading
component under the conditions tested here.

% ============================================================
\section{Discussion}
\label{sec:discussion}
% ============================================================

The results support using a trained language model to read the
report in a checker supplied with image-classifier scores. Both
checkers receive the same evidence. Their different accuracies
therefore concern the comparison of that evidence with the text.
The gap is small on the \textsc{Omission} category and larger on
the other item types, with differences of 18 to 33 percentage points.
This pattern is consistent with better interpretation of report
wording, including uncertainty and negation. Because each category
contains several kinds of language, a separate analysis of annotated
phrases would be needed to measure the contribution of each kind.

The practical role of training is clear in this comparison. The
base language model performs poorly without task-specific training,
while the fine-tuned model learns to produce judgments that agree
with the reference answers. Those answers combine states from the
reference labeler with score thresholds. Rad-Linter follows these
conventions more accurately on held-out pairs than the tested rule
baseline, which uses a different report-reading implementation.
This result does not establish an advantage over the procedure
that generates the answers, or over rule-based methods in general.
The agreement labeler helps select the primary reports; its agreement
with the reference labeler does not supply independent clinical truth.

The two-stage design also makes the source of a flag easier to
inspect. The image classifier supplies a score for a finding;
the checker decides whether the report agrees with that score.
Keeping these roles separate lets us hold the image evidence
constant when comparing checkers. It also leaves an interface that
another classifier could supply. Changing the classifier may change
the score distribution, however, so the thresholds and checker
would need to be tested again. Extending the approach to another
image type, such as CT, would require the same verification with
its own reports and evidence.

The intended use is to alert a clinician to a possible report
problem before the report is signed or acted on. The present test
assesses score--report consistency, while this clinical use remains
to be evaluated. A flag would name
the finding and make the relevant classifier score available for
review. To test this use, the next study should use each image
with its own original report, rather than cross-patient pairings.
Radiologists would judge whether sampled flags identify actual
report errors, and a later study would measure whether showing the
flags helps clinicians correct those errors. The present study
establishes the controlled comparison needed before that step.
Its labels, data coverage, and experimental history define the
limits of what can be concluded now.

% ============================================================
\section{Limitations}
\label{sec:limitations}
% ============================================================

\paragraph{The reference answers are automatic.}
The labels measure agreement with classifier scores on constructed
pairs. Neither the reference labeler nor the agreement labeler was
validated against clinician judgments, and the agreement labeler
was instructed to use the reference labeler's conventions.
Agreement between them therefore does not make
the labels a clinical ground truth. We did not compare alternative
report labelers such as CheXbert~\citep{Smit2020chexbert} or
RadGraph~\citep{Jain2021radgraph}. The image classifier's training
data also include OpenI, so this study does not test its performance
on an entirely new image collection. A checker may correctly follow
a supplied score even when that score is clinically wrong.

\paragraph{The test covers selected cases.}
The primary set has 131 quads in 28 groups linked by shared inputs.
Of these, 99 quads concern Cardiomegaly or Atelectasis. Four
pathologies have fewer than ten quads, \textsc{Hedging} has only
17 items, and Pneumothorax has none. The main results therefore do
not give equally strong evidence for all nine findings. The
agreement filter selects reports that the two labelers interpret
similarly. For the target pathology, the test also uses only high
or low scores, leaving intermediate scores untested. Patients are
separated between training and test, but identical wording across
those splits was not checked: the 3{,}599 reports contain only
2{,}842 distinct texts. Patient separation alone does not rule out
repeated report wording.

\paragraph{The comparison and training choices are limited.}
We test one model size and one rule baseline. Larger zero-shot
models or stronger rule systems may give different results. The
coarse probe was trained on an earlier set of 303 quads, and no
confidence interval was calculated for it. Its role is to test the
specified broad clues, rather than to rule out every possible
shortcut. We also changed the training recipe after the initial
recipe had been evaluated once on both test sets. That initial
recipe reached 51 to 59\% item accuracy on the primary test,
below the rule baseline. Revised recipes were compared on a
development split, while the test sets and rule baseline stayed
fixed. The final result is therefore not an evaluation in which
all training choices were fixed before any test result was seen.
The shared input prompt also requests five output fields, while the
selected fine-tuning targets contain three. The training comparison
therefore reflects the implemented prompt and target setup; its exact
instructions are retained in Appendix~\ref{supp:training}.

\paragraph{Clinical benefit remains to be measured.}
The study uses one imaging modality and one report collection.
It does not measure whether clinicians make fewer errors or whether
patient care improves. Chest X-ray classifiers can miss findings
more often in some patient groups~\citep{SeyedKalantari2021bias};
those errors could affect the checker's flags. We did not evaluate
performance separately for patient subgroups. Clinical evaluation
must assess the image evidence and the checker together. The
supported conclusion is the comparison between the two checkers
under the conditions tested here.

% ============================================================
\Needspace{17\baselineskip}
\section{Conclusion}
\label{sec:conclusion}
% ============================================================

We developed \radlinter{} to help identify potential errors in
chest X-ray reports by comparing the text with image-classifier
scores. The language model is trained to interpret how each finding
is described and decide whether the report and scores agree.
Using the same evidence as the rule baseline, it correctly classifies
98.5\% of pairs as consistent or inconsistent, compared with 80.5\%
for that baseline in the primary controlled test. The advantage remains when reports are included
without requiring agreement between automatic labelers. Training
on 1{,}816 examples is important for the small language model's
performance on this task.

The results support a trained language model as the report-reading
component of this checker. They motivate the next test: applying
it to images with their own reports, asking radiologists to judge
the flags, and measuring whether the flags help clinicians correct
report errors.

% ============================================================
%  BACK MATTER (data, code, compute, funding, AI use)
% ============================================================

\section*{Data and Code Availability}
All experiments use the public, de-identified Indiana University
Chest X-Ray Collection (OpenI)~\citep{Demner2016openi}; no new
patient data were collected and no institutional review was
required.
The release will include the quad construction code and the OpenI
report and image identifiers for every quad. It will also include
the reference labeler and rule baseline with their keyword and cue lists,
the model prompts, the training and evaluation code with its
analysis-plan history, and the trained LoRA adapters. These materials
will be released at \url{https://github.com/Nickcp39/rad_linter}.

\section*{Compute Resources}
All experiments ran on one server with eight NVIDIA RTX A6000 GPUs
(48~GB each).
Each LoRA training run used one GPU. The six final runs, covering
the scores-and-report model (\sysM{}) and report-only model
(\sysT{}) with three seeds each, ran in parallel. Each Mistral
labeling worker used two GPUs.
Software: PyTorch 2.10, Transformers 5.2, PEFT 0.18, and
TorchXRayVision 1.4.0.

\section*{Funding and Competing Interests}
This work received no external funding.
The authors declare no competing interests.

\section*{Correspondence}
Yanda Cheng (\texttt{yc2675@cornell.edu}).

\section*{Acknowledgments}
Claude models (Anthropic) were used for drafting and language
editing, for writing and reviewing the analysis code, and for
cross-checking the reported values against the pipeline outputs.
Codex (OpenAI) was used for subsequent structural and language editing.

% ============================================================
%  REFERENCES
% ============================================================
\begingroup
\setlength{\bibsep}{6.5pt}
\bibliography{refs}
\endgroup

\clearpage
\appendix
\section*{Supplementary Material}
\addcontentsline{toc}{section}{Supplementary Material}
\section{Task Definitions and Label Mapping}
\label{supp:task}

The task compares a report with classifier scores for nine findings:
Atelectasis, Consolidation, Pneumothorax, Edema, Effusion, Pneumonia,
Cardiomegaly, Nodule, and Mass. Write the report as $r$, the radiograph
as $x$, and the score vector as $\mathbf{s}=(s_1,\ldots,s_9)$.
For each finding, the reference labeler distinguishes an affirmation,
an uncertain statement, an explicit negation, and no mention.
For example, ``effusion is present,'' ``possible effusion,'' and
``no effusion'' illustrate the first three states. These are
illustrative statements, not additional clinical test cases.

The reference answer depends on the combination of report state
and score, not on the wording alone. A high score is
$s_p\geq\theta^+=0.6$ and a low score is $s_p\leq\theta^-=0.2$
for finding $p$. Table~\ref{tab:label_mapping} gives the full mapping.
Scores strictly between these cut-offs do not trigger an
inconsistency for that finding. The classifier rescales its outputs
so that $0.5$ is its own decision threshold; these scores are not
clinical probabilities validated in this study.

\begin{table}[!htb]
\centering
\caption{Task labels for one finding. A dash means that this finding
does not trigger a flag. The whole pair receives \textsc{No\_Error}
only when none of the nine findings triggers a flag.}
\label{tab:label_mapping}
\begin{tabular}{lll}
\toprule
Report state & High score ($\geq0.6$) & Low score ($\leq0.2$) \\
\midrule
Affirmed & --- & \textsc{Contradiction} \\
Uncertain & \textsc{Hedging} & --- \\
Explicitly negated & \textsc{Omission} & --- \\
Not mentioned & \textsc{Omission} & --- \\
\bottomrule
\end{tabular}
\end{table}

The label names identify experimental categories. \textsc{Omission}
groups two distinct states: a finding that is not mentioned and
one that is explicitly negated, each paired with a high score.
\textsc{Hedging} applies to uncertain wording only when paired with
a high score. The same uncertain wording paired with a low score
does not trigger a flag under this convention. This is not a
clinical judgment that expressing uncertainty is inappropriate.
\textsc{Contradiction} names the other direction of disagreement:
an affirmed finding paired with a low score. A high score paired
with a negation also disagrees, but belongs to \textsc{Omission}.
\textsc{No\_Error} does not certify that a report is clinically correct.

Each test pair is constructed to have at most one inconsistent
finding. The checker returns three fields: \texttt{has\_error}
states whether it identifies an inconsistency; \texttt{error\_type}
names one of the four labels; and \texttt{finding} names the
affected pathology, or is null when none is flagged. For example,
a high effusion score with a report that never mentions effusion
has an \textsc{Omission} reference label naming Effusion.

The primary test has 114 \textsc{Omission}, 17 \textsc{Hedging},
131 \textsc{Contradiction}, and 262 consistent items. Of the
\textsc{Omission} items, 58 have an unmentioned target finding and
56 have a negated one. Appendix~\ref{supp:evaluation} explains the
five plotted groups; Appendix~\ref{supp:labeling} explains the
three-way state comparison used for agreement filtering.

\clearpage
\section{Reference Labeling, Agreement Filtering, and the Rule Baseline}
\label{supp:labeling}

The \emph{reference labeler} and the \emph{rule baseline} use
different report-reading implementations. The former supplies report
states for the reference answers; the latter is a checker evaluated
against those answers. The \emph{agreement labeler} supplies a second
reading of each report for selecting the primary test. It does not
grade either checker's output.

\subsection{Reference labeler}
The reference labeler is the \texttt{textlib} procedure. It divides
the report into sentences and matches pathology keywords. Its negation
rules extend across a clause, following clinical negation methods
such as NegEx and NegBio~\citep{Chapman2001negex,Peng2018negbio}.
Semicolons and contrast words can end the scope of a preceding
negation, as can an intervening assertion. The procedure also checks
post-keyword negation and expressions such as ``lungs are clear.''
Within a sentence, an applicable negation takes priority over an
uncertainty cue. Across sentences, an affirmation takes priority over
uncertainty, which takes priority over negation; if none is found,
the state is not mentioned. Combining the resulting states with
Appendix~\ref{supp:task}'s thresholds produces the reference answer.
This procedure supplies training and evaluation labels.

\subsection{Rule baseline}
The rule baseline follows a local cue-based approach. It searches
for each pathology keyword and examines up to 50 characters before
and after each match for negation or uncertainty. A match for
``effusion,'' for example, can change meaning when ``no'' occurs
in the relevant context. The keyword and negation lists come from
an earlier rule baseline; uncertainty cues were added for this task.
The baseline uses the same score thresholds and output fields as
Rad-Linter. A high score with an absent or negated match produces
\textsc{Omission}; a high score with a hedged match produces
\textsc{Hedging}; and a low score with an unnegated, unhedged match
produces \textsc{Contradiction}. If several findings trigger flags,
the implementation selects the candidate with the largest distance
beyond its threshold. No candidate yields \textsc{No\_Error}.

\subsection{Agreement labeler and report selection}
The agreement labeler is Mistral-Small-24B-Instruct-2501, a larger
language model~\citep{mistral2025small}. It receives the report alone
and is prompted to use the same report-state conventions. The parser
requires a recognized state for each of the nine findings; an invalid
response cannot qualify a report for the agreement filter.
For this filter, negated and not-mentioned states are merged into
one ``not affirmed'' state. The other two states are affirmed and
uncertain. Agreement is required on all nine of these three-way
states. This is a report-selection condition, followed by the quad
construction requirements; it does not make every agreeing report
part of a test quad.

Agreement on the three-way states ranges from 91.5 to 99.8\% for
individual findings. The labelers agree on all nine in 74.8\% of
reports, and in 53.7\% of the 914 reports that affirm or hedge at
least one finding. The lower all-nine rate requires all states to
agree at once. Agreement does not establish clinical correctness,
especially because both procedures follow the same conventions.
The final training set and the second test do not require this
agreement; the primary test does.

\subsection{Keyword lists and prompt}
The lists below are taken from the two implementations. Complete
cue patterns and matching code are retained in the accompanying
\texttt{reproducibility} directory, with source-file hashes. They
allow exact inspection without treating the two rule systems as
interchangeable.
\paragraph{Reference labeler keywords.}
\noindent\textit{Atelectasis}: atelectasis, atelectatic, volume loss, collapse.\par
\noindent\textit{Consolidation}: consolidation, consolidations, consolidative.\par
\noindent\textit{Pneumothorax}: pneumothorax, pneumothoraces.\par
\noindent\textit{Edema}: edema, vascular congestion, pulmonary congestion.\par
\noindent\textit{Effusion}: effusion, effusions, pleural fluid.\par
\noindent\textit{Pneumonia}: pneumonia, infiltrate, infiltrates, airspace disease, airspace opacity, airspace opacities.\par
\noindent\textit{Cardiomegaly}: cardiomegaly, enlarged heart, cardiac enlargement, enlarged cardiac silhouette, heart is enlarged, heart size is enlarged, heart is mildly enlarged, heart size is mildly enlarged.\par
\noindent\textit{Nodule}: nodule, nodules, nodular.\par
\noindent\textit{Mass}: mass, masses.\par
\paragraph{Rule baseline keywords.}
\noindent\textit{Effusion}: effusion, pleural effusion, pleural fluid, fluid, hydrothorax, pleural fluid collection.\par
\noindent\textit{Pneumonia}: pneumonia, airspace opacity, consolidation, pneumonic, pneumonic infiltrate, airspace disease.\par
\noindent\textit{Pneumothorax}: pneumothorax, collapsed lung, pneumothoraces, air in pleural space.\par
\noindent\textit{Atelectasis}: atelectasis, atelectatic, collapse, collapsed, lung collapse.\par
\noindent\textit{Edema}: edema, oedema, pulmonary edema, fluid overload, pulmonary congestion, interstitial edema.\par
\noindent\textit{Consolidation}: consolidation, consolidated, lobar consolidation, airspace consolidation.\par
\noindent\textit{Mass}: mass, masses, mass-like, density, mass lesion, opacity, mass-like opacity.\par
\noindent\textit{Nodule}: nodule, nodules, nodular, pulmonary nodule, lung nodule, nodular opacity.\par
\noindent\textit{Cardiomegaly}: cardiomegaly, enlarged heart, heart size enlarged, cardiac enlargement, enlarged cardiac silhouette.\par

The agreement labeler uses the following system message and report
prompt. The system message retains its original reference to image
evidence, but this report-labeling call supplies no image scores.
The report placeholder is filled with the unchanged report.
Inference uses temperature 0 and at most 200 new tokens.
\begin{quote}\small
\textbf{System.} You are an expert radiologist auditing chest X-ray reports against image-classifier evidence. Answer with JSON only.

\textbf{User.} Read the chest X-ray report below. For EACH of these 9 pathologies, say how the report describes it:
"affirmed" (stated as present), "negated" (stated as absent, or described as normal), "hedged" (mentioned only with
uncertain language such as possible / may represent / cannot be excluded / questionable), or "not\_mentioned".
Treat synonyms as the same pathology (e.g. "enlarged heart" = Cardiomegaly; "heart size normal" negates Cardiomegaly;
"lungs are clear" negates Pneumonia, Consolidation, Edema and Atelectasis; "infiltrate" or "airspace disease" = Pneumonia;
"pleural fluid" = Effusion; "collapse" or "volume loss" = Atelectasis).

Pathologies: Atelectasis, Consolidation, Pneumothorax, Edema, Effusion, Pneumonia, Cardiomegaly, Nodule, Mass

Report:
[REPORT TEXT]

Return JSON only: \{"Atelectasis": "...", "Consolidation": "...", "Pneumothorax": "...", "Edema": "...", "Effusion": "...", "Pneumonia": "...", "Cardiomegaly": "...", "Nodule": "...", "Mass": "..."\}
\end{quote}

\clearpage
\section{Evaluation Details and Complete Results}
\label{supp:evaluation}

This appendix retains the complete system and result tables, including
the auxiliary comparisons. The identifiers used in the implementation
are \sysM{} (Rad-Linter), \bone{} (rule baseline), \sysZ{} (zero-shot
model with both inputs), \gprobe{} (coarse probe), \sysT{} (trained
report-only model), \btwo{} (zero-shot report-only model), and \sysS{}
(scores-only logistic regression). The reference labeler and agreement
labeler prepare the data; they are not additional rows in these tables.
The primary test is stored as \texttt{test\_main}, and the test
without agreement filtering as \texttt{test\_all}.

\paragraph{Earlier evaluation.}
An earlier evaluation used reports with inserted errors. A model
receiving only the report achieved an F1 score of 90.5; F1 combines
how often a flag is correct with how many target errors are found.
A scores-only probe achieved 67\% accuracy. These results motivated
the present controlled pairings, because they did not establish
that the systems compared report text with image evidence.

\subsection{Construction and dataset composition}
The patient split uses 70\% for training and 30\% for testing.

Two constraints keep the comparison focused. Both vectors must
agree with both reports on the other eight pathologies, leaving
only the target pathology able to disagree. We also limit reuse:
each report $A$ and each low-score vector appears in one quad only.
A high-score vector may appear in up to three quads. The cap rises
to five if the lower cap yields fewer than 120 primary test quads,
as occurred here. Shared inputs are handled together when estimating
uncertainty, as described in \S\ref{sec:method:eval}.

\paragraph{Testing a remaining shortcut.}
A checker might combine the two inputs without comparing specific
findings. If $A$ affirms nothing, the combination of ``nothing
reported as present'' and ``some score is high'' can identify a
problem without identifying the pathology. To make this shortcut
less useful, the construction prefers an $A$ report that affirms
another finding for half of the omission-candidate searches.
In the resulting primary set, 64 of 131 quads have such an $A$
report. We call them \emph{hard quads}, because the checker must go
beyond the simple absence-of-any-finding cue. The remaining quads
are called easy quads. These names describe the available shortcut,
not the clinical difficulty of the cases.

\paragraph{The resulting training and test sets.}
Training uses 454 quads, or 1{,}816 items, with \texttt{textlib}
labels and no agreement filter. The primary test,
\texttt{test\_main}, contains 131 quads, or 524 items, from held-out
reports on which both labelers agree. There are 114
\textsc{Omission}, 17 \textsc{Hedging}, 131 \textsc{Contradiction},
and 262 consistent items. Of the 114 \textsc{Omission} items,
58 have an unmentioned target finding and 56 have a negated one.
The tested pathologies are Cardiomegaly (55 quads), Atelectasis (44),
Effusion (16), Pneumonia (6), Nodule (5), Mass (3), and
Consolidation (2).

We also build \texttt{test\_all} from the same held-out patients
without requiring labeler agreement. It has 181 quads, or 724 items,
with answers from \texttt{textlib} alone. It is constructed
separately, rather than by simply adding cases to \texttt{test\_main}.
Its composition differs, including 40 \textsc{Hedging} quads instead
of 17. No Pneumothorax quad could be formed in either set, and no
Edema quad remained after the agreement filter. This second set
answers a different question from the hard-quad subset: does the
comparison still favor the trained model when reports are not
selected for agreement between labelers?

\subsection{Additional Comparisons and Performance Measures}
\label{supp:metrics}

The main comparison is between \radlinter{} and the rule baseline.
We add five additional comparison systems to check possible explanations for
the results (Table~\ref{tab:systems_complete}). Three receive only one input:
a language model trained on reports alone, the base language model
prompted on reports alone without further training, and logistic
regression, a simple statistical classifier, trained on the scores
alone. All should score 50\% item accuracy and 0\% quad accuracy
on the balanced pairs. These are checks of the test construction.

A fourth system receives both inputs but uses the base language
model without task-specific training. We call this the
\emph{zero-shot} model. It tests how much can be achieved from the
task description alone. The fifth, the \emph{coarse probe}, tests
whether a few broad clues explain success. It is a gradient-boosted
classifier, which combines decision trees, with seven input
features: whether any finding is affirmed, the number affirmed,
whether any is hedged, the maximum and minimum scores, and the
numbers above $\theta^{+}$ and below $\theta^{-}$. The report
features come from the reference labeler. This probe sees summaries of
both inputs but not which score belongs to which reported finding.
Its performance on hard quads provides a comparison for the trained
model's use of more detailed information.

\begin{table}[!htb]
\centering
\small
\setlength{\tabcolsep}{4pt}
\caption{The systems we compare. Above the line, the two checkers;
below it, the reference systems. ``Trained'' means fitted on the
training quads.}
\label{tab:systems_complete}
\begin{tabular}{llll}
\toprule
\textbf{Row} & \textbf{Input} & \textbf{Engine} & \textbf{Trained} \\
\midrule
\sysM{}, \radlinter{} & scores + report & Qwen2.5-1.5B + LoRA & yes \\
\bone{}, rule baseline & scores + report & keyword + negation rules & no \\
\midrule
\sysZ{}, zero-shot & scores + report & Qwen2.5-1.5B, no training & no \\
\gprobe{}, coarse probe & scores + rule-labeler states & gradient boosting, 7 features & yes \\
\sysT{} & report only & Qwen2.5-1.5B + LoRA & yes \\
\btwo{} & report only & Qwen2.5-1.5B, no training & no \\
\sysS{} & scores only & logistic regression & yes \\
\bottomrule
\end{tabular}
\end{table}

The main measures separate individual decisions from consistency
across the four pairings. \emph{Item accuracy} is the fraction of
pairs correctly classified as consistent or inconsistent.
\emph{Quad accuracy} is the fraction of quads for which all four
binary decisions are correct. Correctly answering three pairs in
a quad contributes to item accuracy but does not count as a correct
quad. A single-input checker therefore scores 0\% on quads, while
four independent fair binary guesses are all correct with
probability $(1/2)^4=6.25\%$.

We also measure missed inconsistencies, false flags, and the details
of each prediction. \emph{Sensitivity} is the fraction of
inconsistent items flagged; \emph{specificity} is the fraction of
consistent items left unflagged. \emph{Finding accuracy} checks the
predicted pathology on inconsistent items. \emph{Four-class
accuracy} checks the predicted error-type field against the four
labels in \S\ref{sec:problem}, including \textsc{No\_Error}.
These last two fields are scored separately, not as a joint measure.
The item-type breakdown separates the three inconsistent categories
and the two kinds of consistent pair, giving five groups in the
plot: high score with an affirmed report, and low score with a
report that does not affirm the finding, are shown separately.

To estimate uncertainty, we use a \emph{cluster bootstrap}, which
repeatedly resamples groups of related quads. Quads that share a
report or score vector are related; chains of such shared inputs
form connected groups. There are 28 groups in \texttt{test\_main}
and 34 in \texttt{test\_all}. We resample whole groups 1{,}000 times
and calculate accuracy each time. This avoids treating reused
inputs as independent observations. The middle 95\% of these
estimates gives the reported interval. We use the same resampled
groups for each system, so each difference compares performance
on the same cases.

For repeated training runs, tables report the mean and standard
deviation, which describe the average score and its variation
across seeds. Square brackets show the bootstrap interval where
reported. These measures let us distinguish three questions in
the results: whether the checker detects inconsistencies using the
evidence, whether it improves on rules, and whether that comparison
holds without the labeler-agreement filter.

\paragraph{Interpretation of single-input results.}
The exact 50\% item and 0\% quad result follows when a single-input
checker gives the same valid binary decision for repeated inputs.
It is a consequence of the pairing structure, not evidence that
the checker has no useful language or image knowledge. Parsing
failures are still counted as incorrect and can reduce item accuracy.

\paragraph{Scope of the coarse probe.}
The coarse probe was fitted on an earlier set of 303 quads, rather
than the final 454-quad language-model training set. No confidence
interval was calculated for it. It tests the specified broad
features, not all possible shortcuts. Its reported comparison is
on the primary test; no result for this probe on the second test
is added here. The hard/easy plot is retained in this appendix,
while the main text reports the key hard-subset comparison.

\subsection{Detecting Inconsistencies Using the Supplied Evidence}
\label{supp:results_primary}

\radlinter{} correctly identifies most inconsistencies in the
primary test. On \texttt{test\_main}, which contains 524 items in
131 quads, it achieves 98.5\% item accuracy and 94.4\% quad accuracy
averaged over three training runs (Table~\ref{tab:primary_complete}). Thus,
almost all individual decisions are correct, and all four decisions
are correct in most quads. The runs differ by at most two correct
item predictions. Of the 262 inconsistent items, the model flags
97.7\%. Of the 262 consistent items, it falsely flags 0.6\%, or
one to two items per run. High accuracy therefore reflects both
finding inconsistencies and avoiding false flags.

The prediction usually also names the correct finding and category.
Finding accuracy on inconsistent items ranges from 97.3 to 98.1\%
across runs. Four-class accuracy, scored independently from the
finding field, ranges from 98.3 to 98.7\% over all items.
Detection accuracy within each inconsistent category is 97.4\%
for \textsc{Omission}, 98.0\% for \textsc{Hedging}, and 98.0\% for
\textsc{Contradiction}. These figures show that the overall result
is not driven by success on only one category.

\begin{table}[!htb]
\centering
\caption{Main results on \texttt{test\_main} (131 quads, 524
items). Hard items: the 256 items of the 64 quads whose report $A$
affirms another pathology. Trained systems: mean $\pm$ s.d.\ over
three training runs (seeds); s.d.\ denotes standard deviation.
Square brackets give the 95\% cluster-bootstrap interval for
untrained systems.
The coarse probe is a single fitted model. The three single-input
rows are 50\%/0\% by construction; chance quad accuracy is 6.25\%.}
\label{tab:primary_complete}
\small
\setlength{\tabcolsep}{4.5pt}
\begin{tabular}{llccc}
\toprule
\textbf{Row} & \textbf{Input} & \textbf{\makecell{Item\\acc.\ (\%)}} & \textbf{\makecell{Quad\\acc.\ (\%)}} & \textbf{\makecell{Hard\\items (\%)}} \\
\midrule
\textbf{\sysM{}, \radlinter{}} & scores + report & \textbf{98.5 $\pm$ 0.2} & \textbf{94.4 $\pm$ 0.4} & \textbf{98.3} \\
\bone{}, rule baseline & scores + report & 80.5 [75.0, 86.6] & 56.5 [36.7, 64.3] & 78.9 \\
\midrule
\sysZ{}, zero-shot & scores + report & 52.7 [48.3, 53.7] & 0.0 [0.0, 0.0] & 50.8 \\
\gprobe{}, coarse probe & scores + rule-labeler states & 75.6 & 42.7 & 56.3 \\
\sysT{} & report only & 50.0 $\pm$ 0.0 & 0.0 $\pm$ 0.0 & 50.0 \\
\btwo{} & report only & 50.0 & 0.0 & 50.0 \\
\sysS{} & scores only & 50.0 & 0.0 & 50.0 \\
\bottomrule
\end{tabular}
\end{table}

We next examine whether the model could achieve these scores without
using the evidence in the intended way. The three single-input
systems score exactly 50\% item accuracy and 0\% quad accuracy,
confirming the expected behavior of the balanced test. The coarse
probe, which combines broad clues from both inputs, reaches 75.6\%
item accuracy overall. It reaches 94.0\% on the 268 easy items,
where no finding is affirmed in report $A$. On those cases, noticing
``nothing affirmed'' and ``some score is high'' is often enough.
Exceeding 50\% alone therefore does not show that a checker matches
each report statement to the corresponding score.

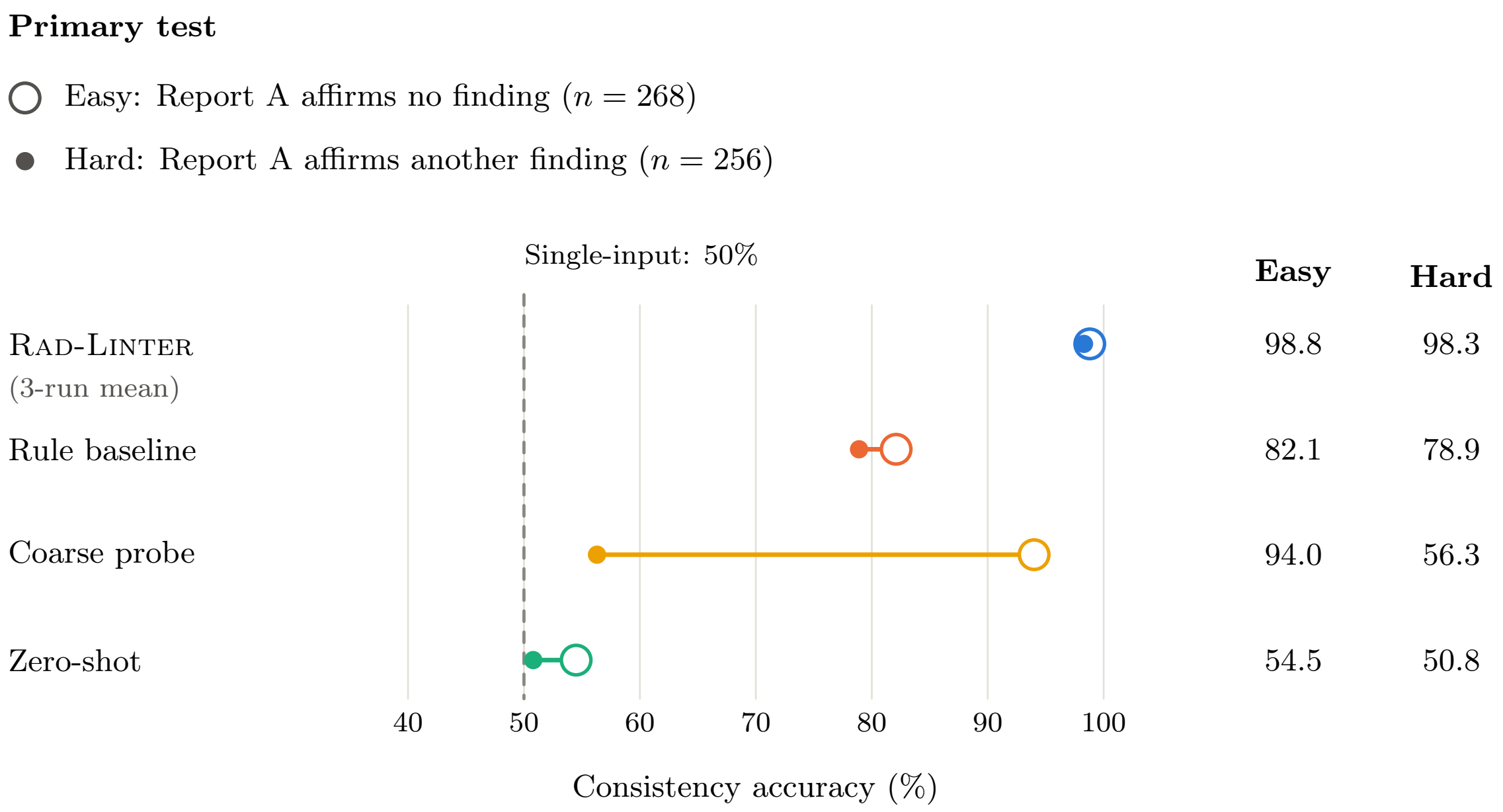
\begin{figure}[!htb]
\centering
% V06: paired point comparison, original 40-100 axis and 50% reference.
\begin{tikzpicture}[fig,x=1pt,y=1pt]
\path[use as bounding box] (0,0) rectangle (432,-230);
\node[anchor=north west,font=\footnotesize\bfseries,inner sep=0pt] at (0,0) {Primary test};
\draw[inkSecondary,fill=white,line width=1pt] (5,-24) circle (4.2pt);
\node[anchor=west,font=\footnotesize,inner sep=0pt] at (16,-24)
 {Easy: Report A affirms no finding ($n=268$)};
\fill[inkSecondary] (5,-42) circle (2.6pt);
\node[anchor=west,font=\footnotesize,inner sep=0pt] at (16,-42)
 {Hard: Report A affirms another finding ($n=256$)};
\def\xo{114}\def\k{3.3}
\foreach \t in {40,50,...,100} {
 \draw[gridGray,line width=.5pt] ({\xo+(\t-40)*\k},-83) -- ({\xo+(\t-40)*\k},-195);
 \node[anchor=north,font=\scriptsize,inner sep=2pt] at ({\xo+(\t-40)*\k},-197) {\t};
}
\draw[inkMuted,dashed,line width=.9pt] (147,-80) -- (147,-195);
\node[anchor=south west,font=\fontsize{8.5}{10}\selectfont,inner sep=0pt] at (147,-73) {Single-input: 50\%};
\node[anchor=south,font=\footnotesize\bfseries,inner sep=0pt] at (366,-78) {Easy};
\node[anchor=south,font=\footnotesize\bfseries,inner sep=0pt] at (411,-78) {Hard};
\foreach \name/\e/\h/\col/\yy in {
 {\radlinter{}}/\fdMEasy/\fdMHard/rlBlue/-94,
 {Rule baseline}/\fdBEasy/\fdBHard/ruleOrange/-124,
 {Coarse probe}/\fdGEasy/\fdGHard/probeYellow/-154,
 {Zero-shot}/\fdZEasy/\fdZHard/zsAqua/-184} {
 \node[anchor=west,font=\footnotesize,inner sep=0pt] at (0,\yy) {\name};
 \draw[\col,line width=1.3pt] ({\xo+(\e-40)*\k},\yy) -- ({\xo+(\h-40)*\k},\yy);
 \draw[\col,fill=white,line width=1pt] ({\xo+(\e-40)*\k},\yy) circle (4.2pt);
 \fill[\col] ({\xo+(\h-40)*\k},\yy) circle (2.6pt);
 \node[font=\footnotesize,inner sep=0pt] at (366,\yy) {\e};
 \node[font=\footnotesize,inner sep=0pt] at (411,\yy) {\h};
}
\node[anchor=west,font=\fontsize{8.5}{10}\selectfont,text=inkSecondary,inner sep=0pt] at (0,-107) {(3-run mean)};
\node[anchor=north,font=\footnotesize,inner sep=0pt] at (213,-216) {Consistency accuracy (\%)};
\end{tikzpicture}
\caption{Binary consistency accuracy on the primary test's easy
and hard subsets. Hollow markers show the 268 easy items, where
report $A$ affirms no finding; filled markers show the 256 hard
items, where it affirms another finding. These groups test a
specific shortcut, not clinical difficulty. The dashed line marks
the 50\% single-input reference. Rad-Linter values average three
training seeds. The coarse probe's easy-item value is derived
from its overall and hard-subset totals.}
\label{fig:hard_easy}
\end{figure}

The hard subset tests beyond that shortcut. Its 256 items come from
64 quads in which $A$ affirms another finding. The coarse probe
falls to 56.3\%, while \radlinter{} changes only from 98.8\% on
easy items to 98.3\% on hard items (Figure~\ref{fig:hard_easy}).
The broad features used by this probe do not explain the model's
performance. The result supports a more detailed comparison of
reported findings and their scores. Having checked this possible
shortcut, we compare the trained model with the rule baseline that
also reads both inputs.

\subsection{Comparing with Rules and Locating the Advantage}
\label{supp:results_rules}

The trained model performs better than the rule baseline in all
three runs. Given the same reports and scores, the rule baseline
achieves 80.5\% item accuracy and 56.5\% quad accuracy, compared
with the model's means of 98.5\% and 94.4\%. These are differences
of about 18 and 38 percentage points, respectively. Here, percentage points measure the difference between the two
accuracy percentages.

The per-run item differences are $+17.7$, $+18.1$, and $+18.1$
points, and the quad differences are $+37.4$, $+38.2$, and $+38.2$.
For each item difference, the paired 95\% bootstrap interval is
$[+13.4,+25.0]$. The quad intervals all lie within
$[+33.1,+63.3]$. All their lower bounds are positive, so the
estimated advantage persists when the related test cases are
resampled together. This measures uncertainty within this test set;
it does not establish performance in a new clinical population.

The model both misses fewer inconsistencies and raises fewer false
flags. Its sensitivity is 97.7\%, compared with 85.5\% for the
rules, while its specificity is 99.4\%, compared with 75.6\%.
The result also holds when the tested pathologies receive equal
weight: mean accuracy is 94.5\% for the model and 81.5\% for the
rules. Individual pathology results vary. The rules range from
65.6\% on Effusion to 91.7\% on Mass; the model ranges from 93.1\%
on Pneumonia to 100\% on Cardiomegaly and Mass, except for 75.0\%
on the eight Consolidation items. The small counts for some findings
limit the precision of those individual estimates.

The item-type breakdown shows where the advantage is larger
(Figure~\ref{fig:by_type}). On \textsc{Omission} items, the rule
baseline and model perform similarly: 98.2\% and 97.4\%,
respectively. This category combines high scores paired with
unmentioned findings and high scores paired with negated findings.
It therefore cannot be read as a separate test of omission or
negation language alone. The rule baseline scores 64.7\% on
\textsc{Hedging}, 77.1\% on \textsc{Contradiction}, and 69.5\%
on consistent pairs in which the report affirms a finding and the
score is high. These cases require identifying how the report
presents the finding, not simply whether a pathology name appears.

The baseline's language rules suggest possible reasons for these
differences. A window extending 50 characters around a keyword may
not connect a distant ``no evidence of'' to that finding. Phrases
such as ``likely atelectasis'' or ``opacity, may represent pneumonia''
require uncertainty to be attached to the right finding. An uncertain
statement may also be followed by an affirmation in another sentence.
These are illustrative language problems, rather than counts from
an independently annotated linguistic analysis. The observed pattern
is consistent with the trained model applying the report-labeling
conventions more reliably. Because both checkers receive the same
scores, the difference arises in how they process the supplied
report and evidence, not from access to a better image classifier.

Training matters for this particular language model. The base model,
which already has general language training but has not been
fine-tuned for this task, reaches only 52.7\% item accuracy and
solves no quad. It flags 224 of 262 inconsistent items but also
210 of 262 consistent ones. In other words, it flags many cases
without reliably separating consistent from inconsistent pairs.
The model trained for 30 epochs on 1{,}816 examples makes that
separation much more accurately. The comparison so far uses reports
selected for agreement between labelers; the second test removes
that selection requirement.

\subsection{Testing Reports Without the Agreement Filter}
\label{supp:results_second}

The model retains its advantage on \texttt{test\_all}, the separately
constructed set of 181 quads that includes reports on which the
labelers disagree. This is a different check from the hard-quad
analysis within \texttt{test\_main}. Here the question is whether
selecting reports with labeler agreement was responsible for the
model's advantage. Reference answers come from \texttt{textlib}
alone, and the case mix changes, with 40 \textsc{Hedging} quads
rather than 17.

\radlinter{} achieves $93.4\pm0.6$\% item accuracy and
$81.4\pm2.1$\% quad accuracy on this set (Table~\ref{tab:second_complete}).
Across runs, item accuracy ranges from 93.0 to 94.1\%, and quad
accuracy from 79.0 to 82.9\%. The rule baseline reaches 75.3\%
and 47.0\%, leaving differences of about 18 percentage points on
items and 34 on quads. The zero-shot model remains near chance,
with 50.8\% item accuracy.

Both checkers score lower than on the primary set. The trained model
loses about five points on items and thirteen on quads. Report
selection, reference labels, and case mix differ between the sets,
so this comparison does not identify a single cause of the decrease.
It does show that the trained model's advantage is not confined to
the reports on which both labelers agree. Together with the primary
comparison and the shortcut checks, this result supports examining
the trained language component as a useful part of the checker.

\begin{table}[!htb]
\centering
\caption{Results on \texttt{test\_all} (181 quads, 724 items; no
agreement filter, labels from the rule labeler alone). Trained
systems: mean $\pm$ standard deviation over three training runs;
untrained systems: 95\% cluster-bootstrap interval in square
brackets, resampling the 34 groups of related quads.}
\label{tab:second_complete}
\begin{tabular}{llcc}
\toprule
\textbf{Row} & \textbf{Input} & \textbf{Item acc.\ (\%)} & \textbf{Quad acc.\ (\%)} \\
\midrule
\textbf{\sysM{}, \radlinter{}} & scores + report & \textbf{93.4 $\pm$ 0.6} & \textbf{81.4 $\pm$ 2.1} \\
\bone{}, rule baseline & scores + report & 75.3 [73.1, 82.9] & 47.0 [37.2, 57.5] \\
\midrule
\sysZ{}, zero-shot & scores + report & 50.8 [50.0, 55.0] & 0.0 [0.0, 0.0] \\
\sysT{} & report only & 50.0 $\pm$ 0.0 & 0.0 $\pm$ 0.0 \\
\btwo{} & report only & 50.0 & 0.0 \\
\sysS{} & scores only & 50.0 & 0.0 \\
\bottomrule
\end{tabular}
\end{table}

\clearpage
\section{Training and Reproducibility}
\label{supp:training}

\subsection{Image classifier and language-model training}
\radlinter{} has two stages: a fixed image classifier supplies the
scores, and a fine-tuned language model reads them with the report
(Figure~\ref{fig:overview}). The classifier is TorchXRayVision's
DenseNet-121~\citep{Cohen2022xrv,Huang2017densenet}, a neural network
for image classification, with the public
\texttt{densenet121-res224-all} weights~\citep{Cohen2020limits}.
Those weights were trained on NIH~\citep{Wang2017chestxray14},
PadChest~\citep{Bustos2020padchest}, CheXpert~\citep{Irvin2019chexpert},
MIMIC-CXR~\citep{Johnson2019mimic}, Google~\citep{Majkowska2020google},
OpenI~\citep{Demner2016openi}, and RSNA~\citep{Shih2019rsna}.
This choice follows work on chest X-ray
classification~\citep{Rajpurkar2017chexnet,Baltruschat2019comparison}
and covers all nine findings needed here. We retain nine of its
18 outputs and do not retrain the classifier.

The scores are rounded to three decimals and written as a JSON
dictionary, a text format that pairs field names with values.
An example is
\texttt{\{"Effusion": 0.823, "Pneumonia": 0.031, \ldots\}}.
The unchanged report and a fixed description of the task follow
these scores in the input. The model returns three fields:
\texttt{has\_error} states whether there is an inconsistency,
\texttt{error\_type} names one of the four labels in Appendix~\ref{supp:task}, and
\texttt{finding} names the affected pathology, or is null if none
is flagged. For instance, a high effusion score with a report that
never mentions effusion should produce an \textsc{Omission} flag
naming Effusion. These examples illustrate the input and output;
they are not additional test cases.

We fine-tune Qwen2.5-1.5B-Instruct~\citep{qwen25} using
LoRA~\citep{hu2022lora}. The 1.5B-parameter model fits on one
consumer GPU, allowing us to test a small language model for this
task. Training uses 1{,}816 labeled report--score pairs, arranged
in 454 quads. Their construction and automatic labels are described
in \S\ref{sec:method:test}. During training, the model learns to
produce the three output fields for each pair. We train for 30
epochs, meaning 30 passes through the training examples, and retain
the final model rather than selecting an earlier saved version.
Only prediction errors in the model's response contribute to the
training loss; the input text is not part of the prediction target.

\paragraph{Training settings.}
LoRA keeps the original model weights fixed and trains small added
modules called adapters. We use rank 16, $\alpha=32$, and dropout
$0.05$ for adapters on the query, key, value, output, gate, up, and
down projections in every model block. We use
AdamW~\citep{Loshchilov2019adamw}, the algorithm that updates the
trainable weights, with a learning rate of $10^{-4}$, a cosine
schedule, and 3\% warm-up. Computation uses bfloat16 precision,
an effective batch size of 32, and a maximum input-and-output length
of 1{,}024 tokens. Tokens are the pieces of text processed by the
model. These settings specify how the comparison model was trained;
the image classifier remains fixed throughout.

An earlier recipe used 10 epochs and five output fields, including
free-text evidence and explanation, but did not fit the training
examples well. We compared revised recipes on a development set
of 45 training quads, reserved for choosing the recipe. The selected
recipe used three output fields, 30 epochs, and training reports
without the labeler-agreement filter described in
\S\ref{sec:method:test}. The initial test exposure is reported in
\S\ref{sec:limitations}. We repeat final training with three random
seeds (42, 43, and 44), which control the randomness of each run.
At test time, every language model selects the highest-probability
next token, called greedy decoding, and generates at most 256 new
tokens. A response that cannot be parsed into the required format
counts as incorrect. The rule baseline provides the comparison described in the main text.

\subsection{Input instructions and output targets}
The input consists of the score dictionary, the report, and the task
instruction; the report-only comparison omits the score dictionary.
The exact instructions below are retained from the implementation.
They ask for five fields, including evidence and explanation, whereas
the selected training recipe supplies targets with only three fields:
\texttt{has\_error}, \texttt{error\_type}, and \texttt{finding}.
This prompt--target difference is part of the reported setup; it has
not been changed for this writing revision. The main item and quad
metrics use the parsed \texttt{has\_error} decision. The parser
accepts surrounding text or Markdown, extracts a JSON object, and
requires that decision field. Unparsable outputs count as incorrect.
Finding and error-type accuracy are scored separately, as described
in Appendix~\ref{supp:evaluation}.
\paragraph{System message.}
\begin{quote}\small
You are a radiology report quality-assurance assistant. Answer with JSON only.
\end{quote}

\paragraph{Task instruction with both inputs.}
\begin{quote}\small
\#\#\# Task:
Compare the radiology report with the visual evidence and decide whether the report contains an error.
Error types:
- OMISSION: a pathology has strong visual evidence (score >= 0.6) but the report does not affirm it.
- HEDGING: a pathology has strong visual evidence (score >= 0.6) but the report only mentions it with uncertain language.
- CONTRADICTION: a pathology has strong evidence against it (score <= 0.2) but the report affirms it.
- NO\_ERROR: none of the above.
Output JSON with fields: has\_error (bool), error\_type, finding (pathology or null), evidence (string or null), explanation (string).
\end{quote}

\paragraph{Task instruction with report only.}
\begin{quote}\small
\#\#\# Task:
Decide whether this chest X-ray report contains an error with respect to the (unseen) image.
Error types: OMISSION (a finding present on the image is not affirmed), HEDGING (a clear finding is only mentioned with
uncertain language), CONTRADICTION (the report affirms a finding the image does not support), NO\_ERROR.
Output JSON with fields: has\_error (bool), error\_type, finding (pathology or null), evidence (string or null), explanation (string).
\end{quote}

\subsection{Recipe history and source records}
The pilot manifest records R3 as the selected recipe, with 1{,}816
final items from the unfiltered training source, rather than the
earlier 303-quad set. The initial test exposure and subsequent
recipe revision are described above and in \S\ref{sec:limitations}.

The accompanying \texttt{reproducibility/SOURCES.json} maps labeling,
construction, training, and evaluation source records to SHA256
hashes. In report selection, the code names \texttt{textlib} and
\texttt{judge} denote the reference labeler and agreement labeler.
Stored code and experimental outputs are unchanged.

\end{document}